\documentclass{article}

\usepackage[preprint]{neurips_2026} 

\usepackage[utf8]{inputenc}
\usepackage[T1]{fontenc}
\usepackage{hyperref}
\usepackage{url}
\usepackage{booktabs}
\usepackage{amsfonts}
\usepackage{amsmath}
\usepackage{caption}
\usepackage{amssymb}
\usepackage{nicefrac}
\usepackage[expansion=false]{microtype}
\usepackage{xcolor}
\usepackage{graphicx}
\usepackage{tikz}
\usetikzlibrary{arrows.meta,positioning,shapes.geometric,calc}
\usepackage{multirow}
\usepackage{array}

\newcolumntype{C}[1]{>{\centering\arraybackslash}p{#1}}

\title{Chainwash: Multi-Step Rewriting Attacks on Diffusion Language Model Watermarks}

\author{
Mohd Ruhul Ameen$^{1}$ \quad
Akif Islam$^{2}$ \quad
Nadim Mahmud$^{3}$ \quad
Md. Ekramul Hamid$^{2}$\\[2mm]
$^{1}$College of Engineering and Computer Science, Marshall University, Huntington, WV, USA\\
$^{2}$Department of Computer Science and Engineering, University of Rajshahi, Rajshahi, Bangladesh\\
$^{3}$Miami University, Oxford, OH 45056, USA\\[1mm]
\texttt{ameen@marshall.edu} \quad
\texttt{s1910776135@ru.ac.bd} \quad
\texttt{mahmudm2@miamioh.edu} \quad
\texttt{ekram\_hamid@ru.ac.bd}
}

\begin{document}

\maketitle

\begin{abstract}
Statistical watermarking is the leading approach for verifying whether text was written by a language model. Most existing schemes assume autoregressive generation, where tokens are produced left to right and contextual hashing is well defined. Diffusion language models (DLMs) generate text by denoising tokens in arbitrary order, so these schemes cannot be applied directly. A recent watermark by Gloaguen et al.\ (ICLR 2026), built on the red--green logit bias framework, addresses this gap and reports true positive detection above 99\% on LLaDA-8B-Instruct outputs. The associated robustness study evaluates the watermark on paraphrased and back-translated text, finds that detectability drops, and shows that the signal can be recovered when the text is long enough. This paper studies what happens when watermarked text is rewritten not once but several times. Using the same watermark configuration on LLaDA-8B-Instruct, 1{,}605 watermarked completions of about 300 tokens each are produced across five WaterBench domains. Each completion is then rewritten by four open-weight language models, from 1.5B to 8B parameters, none of which know the watermark key. Five rewrite styles are tested: paraphrase, humanize, simplify, academic, and summarize-expand. Each rewriter is run on its own, and every style is chained for up to five hops, producing 160{,}500 rewritten texts in total. The reported metrics are detection rate, chainwash success on the originally detected texts, drop in detector signal, and a laundering efficiency score that compares signal loss against semantic preservation. The watermark is detected on 87.9\% of the original 1{,}605 outputs at the standard significance threshold. After a single rewrite, detection falls to between 14\% and 41\% depending on the rewriter and style. After five chained rewrites, detection falls to 4.86\%, meaning 94.76\% of the originally detected texts are no longer flagged. After three rewrites, the detector score has dropped 86\% of the way from its watermarked baseline toward the null distribution. Repeated rewriting is therefore a much stronger attack than a single rewrite, and the result holds across all four rewriters tested.
\end{abstract}

\section{Introduction}
\label{sec:intro}

As language models become widely used for writing, summarization, coding, and
content generation, it becomes increasingly important to know where a piece of
text came from \citep{tang2023science, wu2023survey}. One proposed solution is statistical watermarking: during
generation, the model slightly changes its token choices in a secret,
key-dependent way, so that a detector can later test whether the text was
likely produced by that model. Most of the early and widely studied text
watermarks were designed for autoregressive language models, where tokens are
generated from left to right, one token at a time
\citep{kirchenbauer2023watermark, aaronson2023, kuditipudi2023robust,
dathathri2024synthid}. In that setting, the watermark can use previously
generated tokens as context for deciding which next tokens should carry the
hidden signal.

Diffusion language models (DLMs) break this assumption. Instead of generating
text strictly from left to right, DLMs start from a sequence containing masked
positions and gradually denoise it by filling in tokens, sometimes in arbitrary
order \citep{austin2021d3pm, lou2024sedd, nie2025llada}. When a token is
generated, the ``previous'' context that an autoregressive watermark would
normally use may still be masked, so autoregressive watermarks cannot be
transferred to DLMs directly. Recent work addresses this gap with watermarks
designed for the diffusion setting. \citet{gloaguen2025dlmwm} propose a
DLM-tailored red--green watermark for LLaDA-style generation that applies the
watermark in expectation over possible contexts and biases tokens that help
future tokens become green. The detector remains close to the familiar
red--green binomial test, the generation-time watermark is adapted to
arbitrary-order denoising, and the reported true-positive detection rate
exceeds $99\%$ on fresh DLM outputs. The same study examines robustness under
natural modifications such as paraphrasing and back-translation, and finds
that detectability drops but can be partially recovered when the generated
sequence is long enough.

This paper studies a complementary and more persistent attack setting. In
practice, a user who wants to remove a watermark will rarely edit text by
hand or rewrite it only once. Instead, they can pass the watermarked text
through another instruction-tuned language model and ask for a paraphrase,
a humanized version, a simpler version, a more formal version, or a
summarize-and-expand pass. The rewriter model does not need to know the
watermark key, the original DLM, or the detector. We refer to this setting
as \emph{model-mediated rewriting}. The central question is whether a DLM
watermark survives repeated rewriting by external LLMs while the meaning of
the text is largely preserved.

To answer this question, we evaluate the DLM red--green watermark of
\citet{gloaguen2025dlmwm} under multi-step rewriting attacks. We generate
$1{,}605$ watermarked completions with LLaDA-8B-Instruct across five
WaterBench domains \citep{tu2023waterbench}. Each completion is then rewritten
by four open-weight instruction-tuned rewriter models, ranging from $1.5$B to
$8$B parameters, under five rewrite styles: paraphrase, humanize, simplify,
academic style transfer, and summarize-expand. Each style is chained for up
to five hops, producing $160{,}500$ rewritten texts in total. This setup
allows us to measure not only whether one rewrite weakens the watermark, but
also how watermark detectability changes as the same text is repeatedly
rewritten.

We evaluate the attacks along five axes: the watermark detection rate after
rewriting, the \emph{chainwash} rate (the fraction of originally detected
texts that become undetected), the drop in the detector's continuous signal,
the semantic preservation between original and rewritten text, and a
watermark removal efficiency score that relates watermark-signal loss to semantic
change. Across all four rewriters and five rewrite styles, repeated rewriting
substantially reduces detectability while semantic content is largely
preserved. None of the rewriters has access to the watermark key, the
generator, or the detector. The result is therefore not driven by a privileged
adversary, but by ordinary model-based rewriting that is already widely
available to end users.

This study is intended to extend, not replace, existing DLM watermark
evaluations. Direct detection on fresh outputs and robustness against a
single rewrite remain useful first tests. However, our findings indicate
that they do not by themselves characterize how a watermark behaves once the
text has been rewritten several times by external LLMs. We therefore propose
that iterative, multi-model rewriting should be included as a standard stress
test in future DLM watermark evaluations.

\section{Related Work}
\label{sec:related}

Statistical watermarking for language models was first developed mainly in the
autoregressive setting, where text is generated from left to right. The
red--green watermark of \citet{kirchenbauer2023watermark} partitions the
vocabulary into green and red tokens using a key-dependent hash of the
context, gives green tokens a small logit boost during generation, and
detects the watermark by comparing the number of green tokens against the
expected count under unwatermarked generation. Other autoregressive
watermarking methods explore related ideas with different trade-offs. The
Gumbel-max scheme of \citet{aaronson2023} and the distortion-free construction
of \citet{kuditipudi2023robust} aim to preserve the model distribution more
carefully, while SynthID-Text \citep{dathathri2024synthid} adapts watermarking
for large-scale deployment through tournament sampling.

A parallel line of work studies how robust these watermarks are after the
text is modified. Autoregressive watermarks can be weakened by paraphrasing,
translation, and other text transformations, although the severity of the
attack depends on the watermark, the detector threshold, and the length of
the text \citep{kirchenbauer2024reliability, zhao2023provable,
sadasivan2023ai}. Stronger rewrite attacks, including recursive paraphrasing
and learned paraphrase models, further show that surface-level watermark
signals can be diluted while much of the meaning is preserved
\citep{krishna2023paraphrasing, piet2023markmywords}. These findings motivate
a broader view of watermark robustness: detection on fresh model outputs is
useful, but it does not by itself capture what happens after the text is
rewritten by another model.

The most relevant work for our study is the DLM watermark of
\citet{gloaguen2025dlmwm}. Their method adapts the red--green watermarking
idea to arbitrary-order denoising by applying the watermark in expectation
over possible contexts and biasing tokens that help future tokens carry the
watermark signal. The detector remains close to the familiar red--green
binomial test, while the generation-time watermark is redesigned for the
diffusion setting. Their experiments report true-positive rates above $99\%$
on direct LLaDA-8B generations and evaluate robustness under several natural
modifications, including local edits, paraphrasing, and back-translation.
Other recent works explore watermarking for non-autoregressive or
order-agnostic generation. Order-agnostic schemes such as Unigram
\citep{zhao2023provable} and PatternMark \citep{chen2025patternmark} avoid
depending on a left-to-right context, but introduce separate concerns around
reliability, security, and susceptibility to scrubbing. More recent
DLM-oriented methods, such as Gumbel-style watermarking for discrete
diffusion \citep{chen2025ddlmgumbel} and decoding-order watermarking
approaches such as dgMARK \citep{dgmark2025}, study alternative ways to
embed provenance signals in diffusion generation. The main robustness
evaluations of these schemes still focus primarily on fresh generations,
local text changes, or single-step transformations.

Our work is closest in spirit to the robustness literature, but differs in
the attack setting. Rather than asking only whether a DLM watermark can be
detected on fresh outputs or after one paraphrase, we ask whether it
survives repeated rewriting by external instruction-tuned LLMs. The
adversary does not need access to the watermark key, the watermarked DLM,
or the detector. They only need a public rewriter model and a natural
instruction such as paraphrase, simplify, humanize, or rewrite formally. We
therefore study a multi-hop watermark removal process in which the output of one
rewrite becomes the input to the next. We use WaterBench
\citep{tu2023waterbench} as the source of our prompts because it provides
a recognized multi-domain basis for watermark evaluation, but our focus is
different from the original benchmark: we generate DLM-watermarked outputs
and study how the watermark behaves under repeated rewriting across multiple
rewriter models, rewrite styles, and hops.
\section{Threat Model: Multi-Step Rewriting Attacks}
\label{sec:threat}

We study a black-box rewriting attack against watermarked diffusion language
model outputs. The starting point is a text $x_0$ generated by a watermarked
DLM. A detector can test this text and decide whether it carries the watermark.
The adversary's goal is to produce a new text $x_h$ that preserves the meaning
and usefulness of $x_0$, but is no longer detected as watermarked.

The adversary does not need access to the watermark key, the watermarked DLM,
or the detector. Instead, the adversary only has the generated text and access
to an external instruction-tuned language model used as a rewriter. This
matches a practical setting: a user can copy a watermarked answer into another
LLM and ask for a rewrite using an ordinary natural-language instruction. The
rewriter is therefore key-naive: it does not know that the input text is
watermarked and does not optimize against the detector directly.

Formally, let $x_0$ denote the original watermarked text and let $R$ be a
rewriter model. A one-step rewriting attack produces
\[
x_1 = R(x_0, a),
\]
where $a$ is a rewrite instruction such as paraphrase, simplify, humanize, or
rewrite in an academic style. A multi-step rewriting attack repeats the same
process:
\[
x_h = R(x_{h-1}, a), \qquad h = 1,\dots,H.
\]
We call each rewriting step a \emph{hop}. In our experiments, we evaluate up
to $H=5$ hops. The same rewrite style is kept fixed along a chain so that the
effect of increasing the number of rewrites can be measured directly.

The attack succeeds when the final rewritten text is not detected by the
watermark detector while still preserving the original response meaning. If
$D(\cdot)$ denotes the detector and $p(\cdot)$ its returned $p$-value, then
the original text is detected when $p(x_0)<0.05$. After rewriting, the attack
is successful for that sample if
\[
p(x_0)<0.05 \quad \text{and} \quad p(x_h)\geq 0.05.
\]
This condition captures the core watermark removal event: a text that was originally
flagged as watermarked becomes unflagged after rewriting.

We refer to this process as \emph{chainwash}: the watermark signal is washed
out through a chain of model-mediated rewrites. Unlike token-level deletion,
substitution, or manual editing, chainwash is generative. The output is not
formed by small local changes to the original text, but by repeatedly
regenerating the text through another language model. This makes the attack
different from a single paraphrase: each hop can introduce new lexical choices,
new sentence structure, and new ordering of ideas, while still aiming to keep
the same semantic content.
\section{Experimental Setup}
\label{sec:setup}

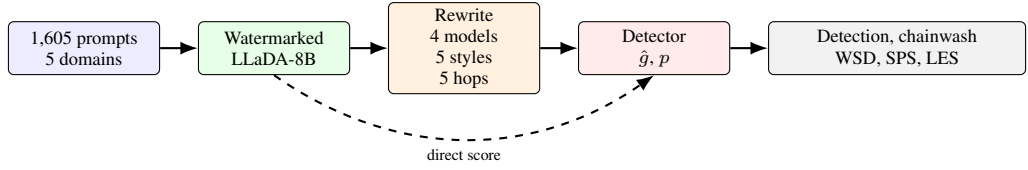
\begin{figure}[t]
\centering
\begin{tikzpicture}[
  node distance=5mm,
  every node/.style={font=\scriptsize},
  box/.style={
    draw,
    rounded corners=2pt,
    minimum height=7mm,
    minimum width=20mm,
    align=center,
    inner sep=2pt
  },
  src/.style={box, fill=blue!7},
  dlm/.style={box, fill=green!10},
  atk/.style={box, fill=orange!12},
  det/.style={box, fill=red!8},
  metric/.style={box, fill=gray!10, minimum width=34mm},
  arr/.style={-Latex, thick},
]

\node[src] (prompts) {1{,}605 prompts\\5 domains};
\node[dlm, right=of prompts] (dlm) {Watermarked\\LLaDA-8B};
\node[atk, right=of dlm] (atk) {Rewrite\\4 models\\5 styles\\5 hops};
\node[det, right=of atk] (det) {Detector\\$\hat g$, $p$};
\node[metric, right=of det] (metrics) {Detection, chainwash\\WSD, SPS, LES};

\draw[arr] (prompts) -- (dlm);
\draw[arr] (dlm) -- (atk);
\draw[arr] (atk) -- (det);
\draw[arr] (det) -- (metrics);

\draw[arr, dashed] (dlm.south) to[bend right=35]
node[below,font=\tiny]{direct score} (det.south);

\end{tikzpicture}

\caption{Experimental pipeline. We generate watermarked DLM outputs from a
fixed WaterBench prompt set, rewrite each output with key-naive external LLMs
across styles and hops, and re-score original and rewritten texts with the same
watermark detector.}
\label{fig:pipeline}
\end{figure}

Our evaluation studies whether a DLM watermark remains detectable after the
watermarked text is repeatedly rewritten by external language models. The
pipeline has four stages. First, we assemble a fixed prompt set from
WaterBench. Second, we generate watermarked responses with LLaDA-8B-Instruct
using the DLM watermark of \citet{gloaguen2025dlmwm}. Third, we rewrite each
watermarked response using external instruction-tuned LLMs under several
rewrite styles and hop counts. Finally, we re-score every original and rewritten
text with the same watermark detector. Figure~\ref{fig:pipeline} summarizes
the full pipeline.

\subsection{Relationship to the Original Watermark Configuration}
\label{sec:setup-relation}

Our setup follows \citet{gloaguen2025dlmwm} closely, but differs in two
intentional ways. First, we use watermark strength $\delta=3$ rather than
$\delta=4$. The original paper reports above $99\%$ true-positive detection
near $300$ tokens at $\delta=4$, while our $\delta=3$ setting gives an
original detection rate of $87.91\%$ on $1{,}605$ LLaDA-8B-Instruct outputs.
This setting evaluates a still-detectable watermark with lower expected
quality distortion. Second, instead of using one closed-source paraphraser, we
use four open-weight rewriters from $1.5$B to $8$B parameters under five
rewrite styles. Therefore, our single-hop numbers should not be read as a
direct comparison to the original robustness results. The contribution is the
multi-hop trajectory: repeated rewriting continues to remove watermark signal
after the first rewrite.

\subsection{Source Prompts}
\label{sec:prompts}

We use $1{,}605$ prompts from five WaterBench subsets
\citep{tu2023waterbench}, covering open-ended explanatory question answering,
personal-finance question answering, multi-document news summarization,
query-based meeting summarization, and general instruction following. This
gives the evaluation a mixture of short prompts, prompts with short context,
and prompts with long source documents. Table~\ref{tab:prompts} reports the
prompt distribution. The same prompt set is used for all rewriter models,
rewrite styles, and hop counts.

\begin{table}[t]
\centering
\caption{Source prompt distribution. ``Avg.\ ctx.'' denotes average context
length in characters.}
\label{tab:prompts}
\small
\begin{tabular}{lrrrl}
\toprule
WaterBench subset & $N$ & Avg.\ input & Avg.\ ctx. & Domain \\
\midrule
\textsc{2-1 longform\_qa}   & 200 & 197 & --      & ELI5 explanatory QA \\
\textsc{2-2 finance\_qa}    & 200 & 67  & --      & Personal-finance QA \\
\textsc{4-1 multi\_news}    & 200 & 0   & 12{,}084 & Multi-document news summarization \\
\textsc{4-2 qmsum}          & 200 & 74  & 57{,}460 & Meeting summarization \\
\textsc{5-1 alpacafarm}     & 805 & 44  & 120     & Open instruction following \\
\midrule
\textbf{Total}              & \textbf{1{,}605} & & & \\
\bottomrule
\end{tabular}
\end{table}

\subsection{Watermarked DLM Generation}
\label{sec:gen}

We generate the initial watermarked texts with LLaDA-8B-Instruct
\citep{nie2025llada}. The watermark is the DLM red--green watermark proposed
by \citet{gloaguen2025dlmwm}. We use SumHash seeding, Bernoulli green-list
assignment with $\gamma=0.25$, green-token bias $\delta=3.0$, and top-$k=50$
candidate selection. For DLM generation, we use $300$ denoising steps,
generation length $300$, block length $25$, sampling temperature $0.5$, and
CFG scale $0$. Generation is run in bfloat16 on a single NVIDIA RTX~5090 GPU.
For each prompt, we save the prompt identifier, source subset, generated
watermarked text, generated length, detector green-list fraction $\hat{g}$,
detector $p$-value, and binary watermark decision at the standard threshold
$p<0.05$. These original detector scores are used as the reference point for
all post-rewrite measurements; in particular, chainwash is computed relative
to the texts that were detected as watermarked before rewriting.

\subsection{Rewriter Models}
\label{sec:rewriters}

We evaluate four open-weight instruction-tuned LLMs as external rewriters:
Gemma-4-E4B-it, Llama-3.1-8B-Instruct \citep{llama2024tech},
Qwen2.5-1.5B-Instruct, and Qwen2.5-7B-Instruct \citep{qwen2025tech}. The
rewriters are key-naive: they are not given the watermark key, the DLM
weights, the watermarking algorithm, or the detector score. Each rewriter
only receives the text to be rewritten and a natural-language rewrite
instruction. Each rewriter is evaluated independently, so a five-hop chain
uses the same rewriter at every hop and we do not mix different rewriter
models inside the same chain. All rewriters are run locally in bfloat16 with
sampling temperature $0.7$, top-$p=0.9$, and a maximum of $360$ new tokens.

\subsection{Rewrite Styles and Hop Chains}
\label{sec:attacks}

We test five rewrite styles that represent common ways a user might ask an
LLM to revise text:

\begin{itemize}
    \item \textbf{Paraphrase:} rewrite the text using substantially different
    wording while preserving the same meaning.
    \item \textbf{Humanize:} rewrite the text so that it sounds more natural,
    human-written, and conversational.
    \item \textbf{Simplify:} rewrite the text in simpler language for a
    general audience.
    \item \textbf{Academic:} rewrite the text in a more formal academic
    style.
    \item \textbf{Summarize-expand:} first compress the content, then expand
    it back into a coherent rewritten response with the same main meaning.
\end{itemize}

For each original watermarked text, each rewriter model, and each rewrite
style, we run a chain of up to five hops. Let $x_0$ denote the original
watermarked output, $M$ denote a rewriter model, and $a$ denote a rewrite
style. The rewritten text at hop $h$ is

\[
x_h = M(x_{h-1}, a), \qquad h=1,\dots,5.
\]

The rewrite style and rewriter model are fixed within a chain. Thus, hop $1$
measures a single rewrite, while hops $2$ through $5$ measure the effect of
repeatedly rewriting the previous output under the same style, which isolates
the role of hop count. The full experiment produces
$1{,}605 \times 4 \times 5 \times 5 = 160{,}500$ rewritten texts. Each
rewritten text is scored with the same detector used for the original DLM
outputs.
\subsection{Evaluation Metrics}
\label{sec:metrics}

We report binary detection, chainwash, continuous signal drop, semantic
preservation, and watermark removal efficiency. Detection rate is the percentage of
rewritten texts still flagged at $p<0.05$:
\[
\mathrm{DetectionRate}(a,h)
=
\frac{1}{N}
\sum_{i=1}^{N}
\mathbf{1}\!\left[p(y^{(i)}_{a,h}) < 0.05\right].
\]
Chainwash measures how many originally detected texts become undetected after
rewriting. Let $\mathcal{D}_0=\{i:p(x^{(i)}_0)<0.05\}$ denote the originally
detected subset. Then
\[
\mathrm{Chainwash}(a,h)
=
\frac{1}{|\mathcal{D}_0|}
\sum_{i\in\mathcal{D}_0}
\mathbf{1}\!\left[p(y^{(i)}_{a,h}) \geq 0.05\right].
\]
We also track watermark signal drop using the detector's green-list fraction
$\hat{g}$, semantic preservation score (SPS) using sentence-embedding cosine
similarity between the original and rewritten text, and watermark removal efficiency
score (LES), which normalizes detector-score drop by semantic change. Full
metric definitions are given in Appendix~\ref{app:metrics}.

\section{Results}
\label{sec:results}

This section evaluates how the watermark behaves after repeated rewriting. We
first verify the direct detection baseline, then study how detection changes
with rewrite hops, rewriter models, attack styles, semantic preservation, and
WaterBench domain. Unless otherwise stated, detection uses the standard
threshold $p<0.05$.

\subsection{Direct Detection Baseline}
\label{sec:results-baseline}

Before rewriting, the watermark is detected in $87.91\%$ of the original
$1{,}605$ LLaDA-8B-Instruct outputs at the standard threshold $p<0.05$. The
mean detector score is $\hat{g}=0.3743$, which is substantially above the null
green-list rate $\gamma=0.25$. This confirms that the watermark is active and
clearly detectable before any rewriting attack is applied.

This direct detection rate is lower than the above-$99\%$ figure reported by
\citet{gloaguen2025dlmwm} at $\delta=4$, and is consistent with our use of the
weaker $\delta=3$ setting (Section~\ref{sec:setup-relation}). This difference
does not affect the main goal of our experiment, which is to measure how the
watermark signal changes as the same watermarked outputs are repeatedly
rewritten.

This baseline is important for interpreting the rest of the results. A
rewriting attack can only remove detection from samples that were detected in
the first place. Therefore, we report both detection rate after rewriting and
chainwash rate. Detection rate measures how many rewritten texts remain flagged,
while chainwash measures how many originally detected texts are no longer
flagged after rewriting.

\subsection{Detection Drops Rapidly Under Repeated Rewriting}
\label{sec:results-detection-decay}

Figure~\ref{fig:detection-chainwash}(a) shows the detection rate after each rewrite
hop, averaged over the five rewrite styles for each rewriter model. The dashed
line shows the original detection rate before rewriting. All four rewriters
cause a large drop after the first hop, and additional hops keep the detection
rate low.

The first rewrite is already strong. Across rewriter models, the average
detection rate after hop~1 falls far below the direct baseline. Gemma-4-E4B
starts with the highest hop-1 detection among the four rewriters, while
Llama-3.1-8B and Qwen2.5-7B are stronger single-hop rewriters. However, after
multiple hops, all models converge to a low-detection region. By hop~5, the
mean detection rate for each rewriter is close to the low-teens percentage
range.

\subsection{Hop-5 Chainwash Depends on Rewriter and Rewrite Style}
\label{sec:results-chainwash-heatmap}

\begin{figure}[t]
\centering

\begin{minipage}[t]{0.49\linewidth}
\centering
\includegraphics[width=\linewidth]{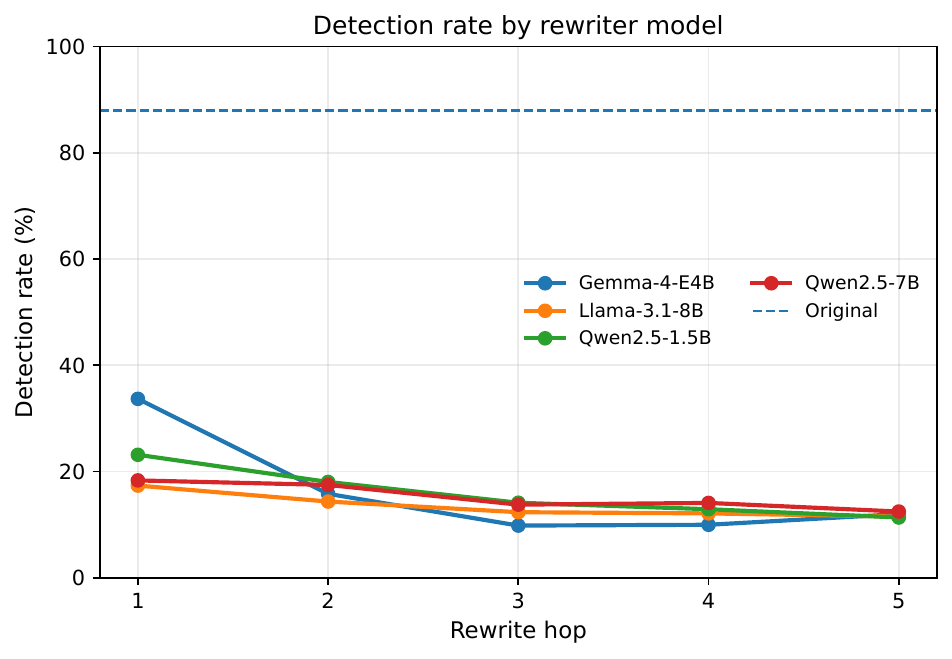}
\caption*{\textbf{(a)} Detection rate by rewriter model across rewrite hops.}
\end{minipage}
\hfill
\begin{minipage}[t]{0.49\linewidth}
\centering
\includegraphics[width=\linewidth]{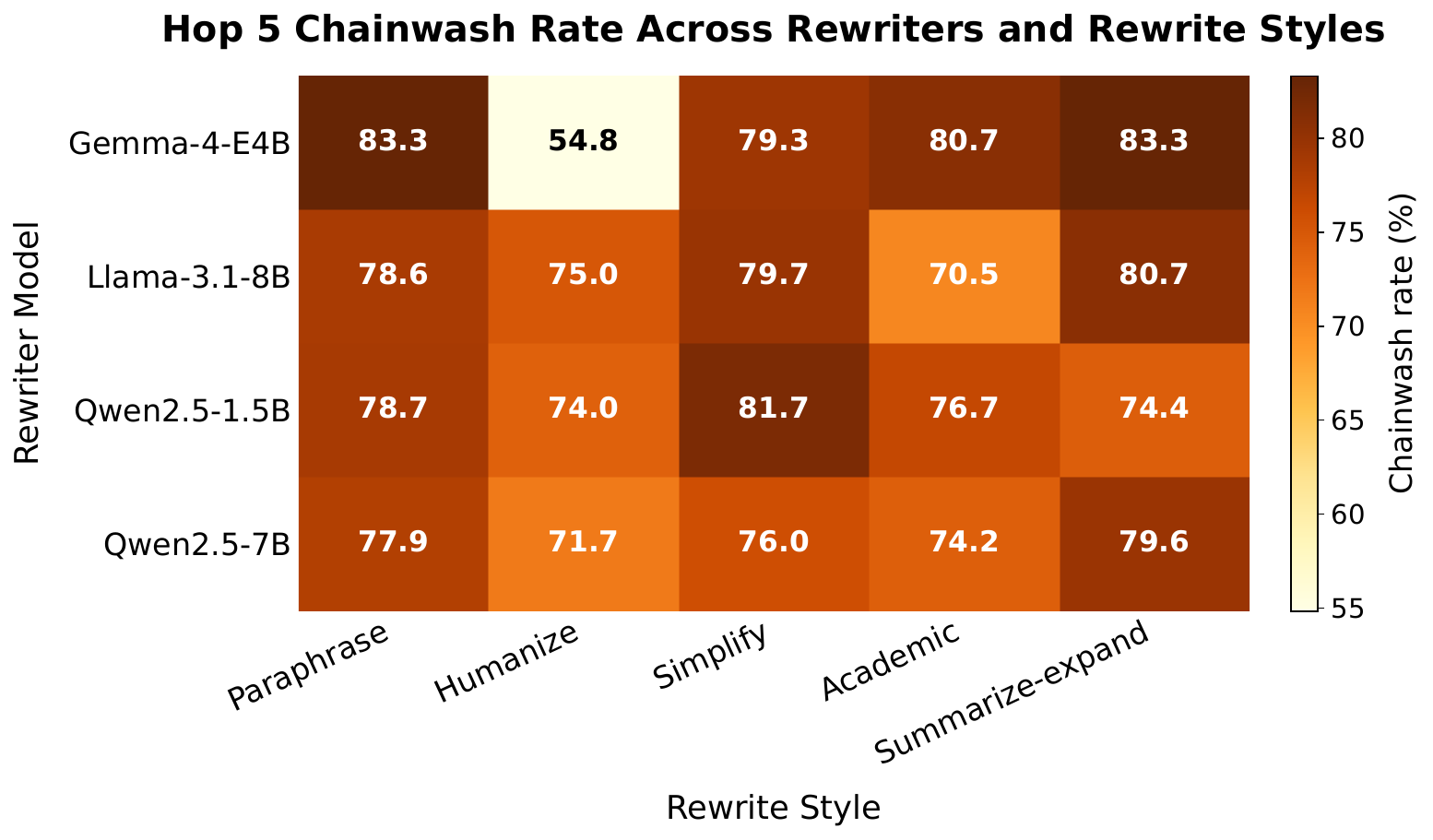}
\caption*{\textbf{(b)} Hop-5 chainwash rate by rewriter and rewrite style.}
\end{minipage}

\caption{Detection decay and hop-5 chainwash under repeated rewriting. 
(a) Detection rate at $p<0.05$ across rewrite hops, averaged over the five
rewrite styles for each rewriter model. The dashed line shows the original
detection rate before rewriting. 
(b) Hop-5 all-output chainwash rate over the full set of $1{,}605$ outputs.
Higher values indicate that more originally detected watermarked texts become
undetected after five rewrite hops.}
\label{fig:detection-chainwash}
\end{figure}

Figure~\ref{fig:detection-chainwash}(b) reports the hop-5 chainwash rate for each
rewriter and rewrite style. This figure uses the all-output chainwash rate,
where the denominator is the full set of $1{,}605$ original outputs. Since the
direct detection baseline is $87.91\%$, an all-output chainwash rate near
$83\%$ corresponds to removing detection from almost all originally detected
samples.

The strongest hop-5 settings are Gemma-4-E4B with paraphrase and
summarize-expand, each reaching an all-output chainwash rate of $83.3\%$.
Because the original detection rate is $87.91\%$, this corresponds to removing
the watermark decision from approximately $94.8\%$ of the originally detected
outputs. This matches the central claim of the paper: repeated rewriting can
wash out the detector decision for most samples that were initially detected as
watermarked.

\begin{table}[t]
\centering
\caption{Hop-5 all-output chainwash rate (\%). Values are computed over all
$1{,}605$ original outputs. The maximum observed rate is $83.3\%$, achieved by
Gemma-4-E4B under paraphrase and summarize-expand.}
\label{tab:hop5-chainwash}
\small
\begin{tabular}{lrrrrr}
\toprule
Rewriter & Paraphrase & Humanize & Simplify & Academic & Summarize-expand \\
\midrule
Gemma-4-E4B  & \textbf{83.3} & 54.8 & 79.3 & 80.7 & \textbf{83.3} \\
Llama-3.1-8B & 78.6 & 75.0 & 79.7 & 70.5 & 80.7 \\
Qwen2.5-1.5B & 78.7 & 74.0 & 81.7 & 76.7 & 74.4 \\
Qwen2.5-7B   & 77.9 & 71.7 & 76.0 & 74.2 & 79.6 \\
\bottomrule
\end{tabular}
\end{table}

Several patterns are visible. First, paraphrase, simplify, academic rewriting,
and summarize-expand are consistently strong across models. Second, humanize is
weaker for Gemma-4-E4B, reaching only $54.8\%$ all-output chainwash at hop~5.
This suggests that not all natural rewriting instructions remove the watermark
signal equally. Third, model size alone does not explain attack strength:
Qwen2.5-1.5B is often competitive with Qwen2.5-7B, and Gemma-4-E4B is the
strongest model for some styles but not all.

\subsection{Meaning Is Largely Preserved Across Hops}
\label{sec:results-sps}

\begin{figure}[t]
\centering

\begin{minipage}[t]{0.49\linewidth}
\centering
\includegraphics[width=\linewidth]{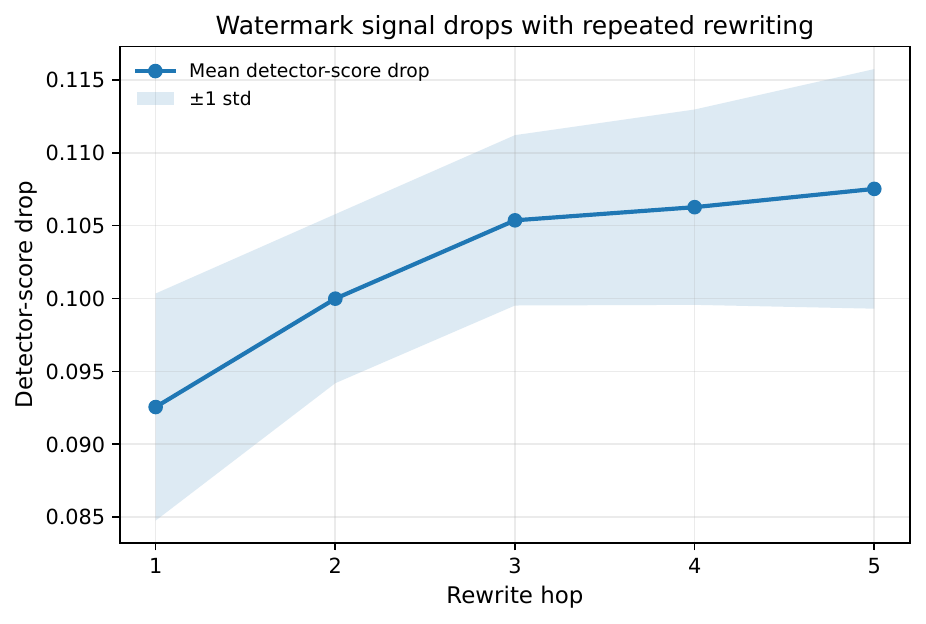}
\caption*{\textbf{(a)} Watermark signal drop across rewrite hops.}
\end{minipage}
\hfill
\begin{minipage}[t]{0.49\linewidth}
\centering
\includegraphics[width=\linewidth]{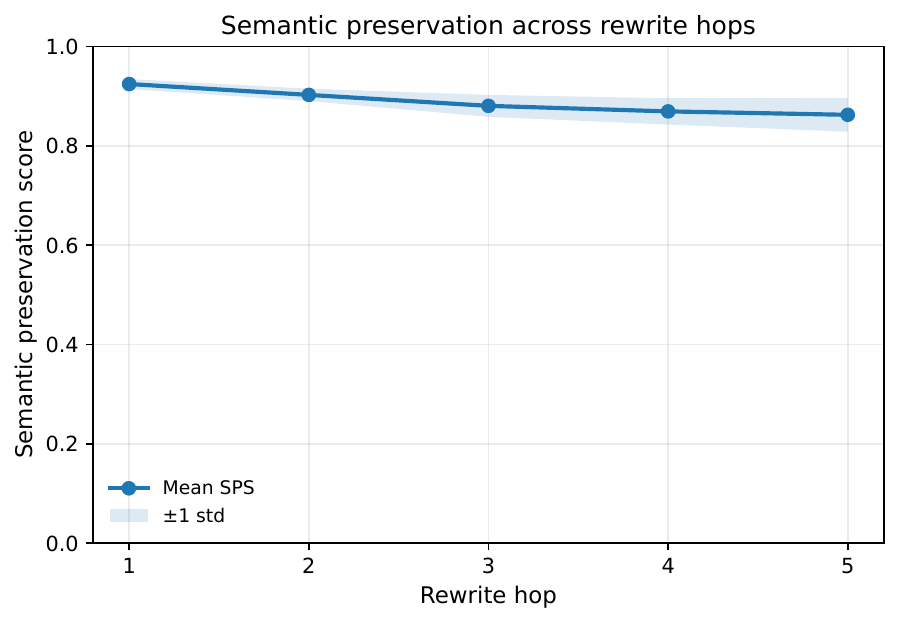}
\caption*{\textbf{(b)} Semantic preservation across rewrite hops.}
\end{minipage}

\caption{Signal removal and semantic preservation under repeated rewriting.
(a) The detector-score drop increases from hop~1 to hop~5, showing that
repeated rewriting removes more of the underlying watermark signal.
(b) Mean semantic preservation score remains high across rewrite hops, with the
shaded region showing one standard deviation. Even after five hops, rewritten
texts remain semantically close to the original watermarked outputs.}
\label{fig:wsd-sps-hop}
\end{figure}

A strong rewriting attack should not simply destroy the text. It should remove
the watermark signal while preserving meaning. Figure~\ref{fig:wsd-sps-hop}(b) shows
the semantic preservation score (SPS) across rewrite hops. SPS is computed
between the original watermarked text and the rewritten text. Higher values
indicate stronger semantic preservation.

The mean SPS remains high throughout the chain. It starts above $0.92$ after
one rewrite and remains around $0.86$ even after five rewrite hops. This means
that the observed detector failure is not caused by completely changing or
destroying the text. The rewritten outputs remain close in meaning to the
original watermarked outputs while the watermark signal is reduced.

\subsection{The Continuous Watermark Signal Drops With Hops}
\label{sec:results-wsd}

Binary detection only shows whether a text is flagged or not. To understand
whether the underlying watermark signal is actually being removed, we measure
the drop in detector score. Figure~\ref{fig:wsd-sps-hop}(a) shows that the mean
detector-score drop increases with rewrite hop.

\begin{table}[t]
\centering
\caption{Mean detector score $\hat{g}$ across rewrite hops, averaged over all rewriters and rewrite styles. The null green-list rate is $\gamma=0.25$. By hop~3, the mean detector score has dropped about $85\%$ of the way from the watermarked baseline toward the null rate.}
\label{tab:ghat-trajectory}
\small
\begin{tabular}{lcccccc}
\toprule
Hop & Baseline & 1 & 2 & 3 & 4 & 5 \\
\midrule
Mean $\hat{g}$ & 0.374 & 0.282 & 0.274 & 0.269 & 0.268 & 0.267 \\
Distance to null & +0.124 & +0.032 & +0.024 & +0.019 & +0.018 & +0.017 \\
Reduction toward null & --- & 74\% & 80\% & 85\% & 86\% & 86\% \\
\bottomrule
\end{tabular}
\end{table}

The detector-score drop grows from $0.093$ at hop~1 to $0.108$ at hop~5. In
absolute terms (Table~\ref{tab:ghat-trajectory}), the mean detector score
$\hat{g}$ falls from the baseline of $0.374$ to $0.282$ after one rewrite,
and continues falling to $0.267$ after five rewrites. The null green-list
rate is $\gamma=0.25$, so by hop~3 the mean detector score has lost
approximately $85\%$ of its initial gap above null. The watermark signal is
therefore not merely weakened but largely removed across the corpus, while
remaining slightly above the null distribution. The largest reduction occurs
between hop~1 and hop~3, after which the curve flattens, suggesting that most
of the watermark signal is removed within the first few rewrites.

Together with the SPS result, this shows a clear trade-off: repeated
rewriting reduces the detector signal while maintaining high semantic
similarity. This is precisely the behavior expected from a practical
rewriting attack. A full scatter plot of detector-score drop against semantic preservation for
all model-style-hop settings is provided in Appendix~\ref{app:pareto}.
The effect also holds across all five WaterBench domains, with domain-level
results reported in Appendix~\ref{app:domain}.
Overall, the direct detector identifies $87.91\%$ of the original outputs.
After five rewrite hops, the strongest settings chainwash $83.3\%$ of all
outputs, corresponding to approximately $94.8\%$ of the originally detected
subset. Mean SPS remains around $0.86$ at hop~5, while the mean detector-score
drop reaches approximately $0.108$.
\section{Discussion}
\label{sec:discussion}

\paragraph{Implications for evaluation.}
Direct detection gives an incomplete picture of DLM watermark robustness. The
watermark is clearly active before rewriting, with $87.91\%$ detection on the
original LLaDA-8B-Instruct outputs. However, after five rewrite hops, the
strongest settings remove the watermark decision from $83.3\%$ of all outputs,
corresponding to approximately $94.8\%$ of the originally detected subset.
This does not mean that the studied watermark is ineffective on fresh
generations. Rather, it shows that direct detection and single-rewrite
robustness do not fully describe deployment behavior. Future DLM watermark
evaluations should include multi-hop rewriting, multiple rewriter models,
multiple rewrite instructions, semantic preservation, and continuous
signal-drop metrics.

\paragraph{What the attack reveals.}
Repeated rewriting is different from a single rewrite because each hop
regenerates the surface form of the text. Most signal loss occurs within the
first few hops, after which detection remains low. Rewriter size alone does
not explain the effect: Qwen2.5-1.5B is often competitive with Qwen2.5-7B, and
Gemma-4-E4B gives some of the strongest hop-5 results. Rewrite style also
matters. Paraphrase, simplify, academic rewriting, and summarize-expand are
strong across models, while humanize is weaker in some settings. Since mean
SPS remains around $0.86$ at hop~5, the watermark is weakened without simply
destroying the original meaning.

\section{Conclusion}
\label{sec:conclusion}

This paper studied whether a recent diffusion language model watermark
remains detectable after repeated model-mediated rewriting. Using the same
watermark configuration on LLaDA-8B-Instruct, we generated $1{,}605$
watermarked outputs of about $300$ tokens and rewrote them using four
key-naive open-weight LLMs, five rewrite styles, and up to five rewrite hops,
producing $160{,}500$ rewritten texts.

The watermark is detected on $87.91\%$ of the original outputs before
rewriting. After repeated rewriting, detection falls sharply. In the
strongest hop-5 settings, $83.3\%$ of all outputs are chainwashed,
corresponding to approximately $94.8\%$ of the originally detected subset. At
the same time, semantic preservation remains high, showing that the watermark
is weakened without simply destroying the original meaning. These results
show that multi-hop rewriting is a stronger and distinct robustness setting
for DLM watermarks. Future watermark evaluations should therefore measure not
only direct detection on fresh generations, but also survival under repeated
rewriting by external LLMs.

\section*{Reproducibility Statement}

We provide the full experimental pipeline needed to reproduce the study:
WaterBench prompt construction, watermarked generation with LLaDA-8B-Instruct,
multi-hop rewriting with open-weight rewriter models, detector re-scoring,
and metric computation. The released artifacts include the original
watermarked generations, rewritten outputs, detector scores, semantic
preservation scores, chainwash summaries, and plotting scripts for all
reported figures and tables. All rewriting experiments are run with fixed
decoding settings for each model and style. The full pipeline is designed to
run on a single high-memory consumer GPU, with the watermarked generation
and rewrite stages executed in bfloat16.

\section*{Ethics Statement}

This work evaluates the robustness of an existing DLM watermark under
realistic rewriting behavior. The attack setting uses ordinary open-weight
language models and natural rewrite instructions, without access to the
watermark key, the detector, or the original DLM internals. The goal is not
to encourage misuse, but to measure a practical weakness that watermark
evaluations should account for. Since watermarking is increasingly discussed
as a tool for content traceability, it is important that robustness claims
reflect how text may be rewritten after generation. We report the results to
support more reliable future watermark design and evaluation. We do not
release tools intended for harmful evasion beyond the research pipeline
needed to reproduce the reported experiments.

\bibliographystyle{plainnat}
\bibliography{references}

\newpage
\appendix

\section{Additional Results}
\label{app:additional-results}

\section{Metric Definitions}
\label{app:metrics}

Watermark signal drop is computed as
\[
\mathrm{WSD}^{\hat{g}}(a,h)
=
\frac{1}{N}
\sum_{i=1}^{N}
\left[
\hat{g}(x^{(i)}_0) - \hat{g}(y^{(i)}_{a,h})
\right].
\]

Semantic preservation score is
\[
\mathrm{SPS}(a,h)
=
\frac{1}{N}
\sum_{i=1}^{N}
\cos
\left(
e(x^{(i)}_0),
e(y^{(i)}_{a,h})
\right),
\]
where $e(\cdot)$ is a sentence embedding model \citep{reimers2019sbert}.

watermark removal efficiency score is
\[
\mathrm{LES}(a,h)
=
\frac{\mathrm{WSD}^{\hat{g}}(a,h)}
{1 - \mathrm{SPS}(a,h) + \epsilon},
\]
with $\epsilon=10^{-3}$.

\subsection{Trade-off Between Semantic Preservation and Watermark Removal}
\label{app:pareto}

\begin{figure}[h]
\centering
\includegraphics[width=0.70\linewidth]{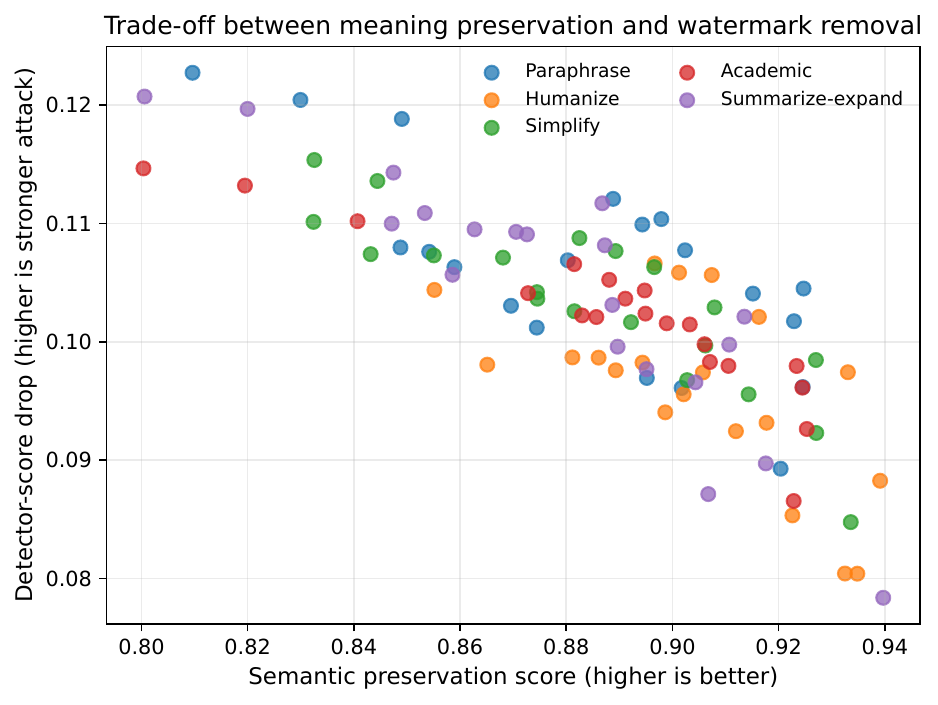}
\caption{Trade-off between semantic preservation and watermark removal. Each
point represents a model-style-hop setting. The $x$-axis shows semantic
preservation, and the $y$-axis shows detector-score drop. Stronger attacks
appear toward the upper-right.}
\label{fig:pareto-appendix}
\end{figure}

Paraphrase and summarize-expand tend to produce larger detector-score drops,
while humanize often preserves meaning well but is not always the strongest at
removing the watermark signal. This confirms that rewrite style affects the
balance between semantic preservation and watermark removal.

\subsection{Domain-Level Chainwash}
\label{app:domain}

\begin{figure}[h]
\centering
\includegraphics[width=0.70\linewidth]{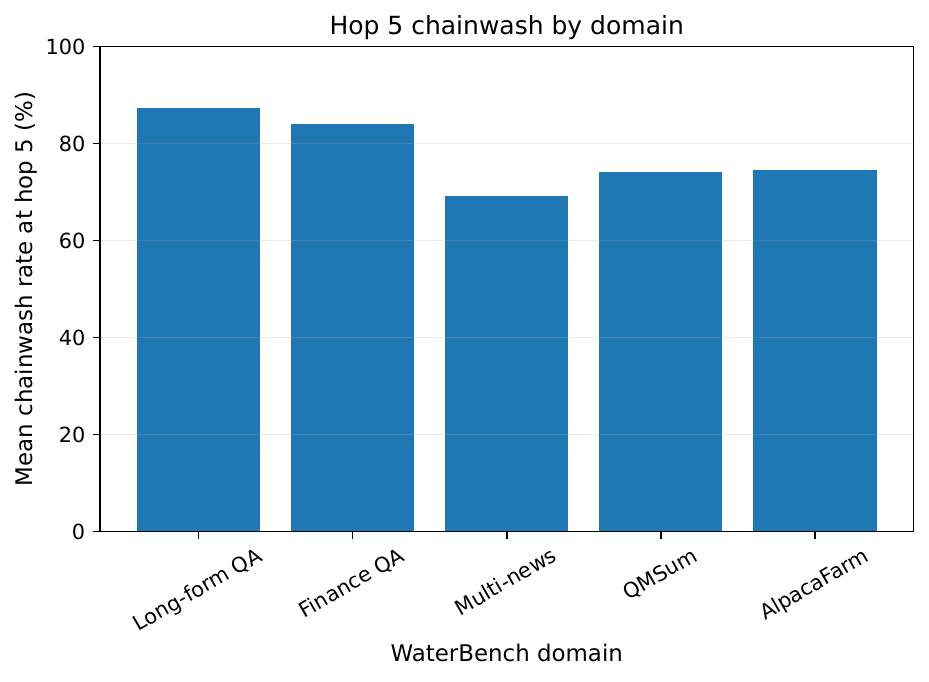}
\caption{Mean hop-5 chainwash rate by WaterBench domain. Chainwash is highest
for long-form question answering and finance question answering, and lower for
multi-document news summarization.}
\label{fig:domain-chainwash-appendix}
\end{figure}

The attack succeeds across all five WaterBench domains, although the rate
varies by task type. Long-form question answering and finance question answering
show the highest hop-5 chainwash rates, while multi-document news summarization
is lower.

\section{Limitations}
\label{app:limitations}

This study evaluates one DLM, LLaDA-8B-Instruct, and one DLM watermarking
method, the red--green watermark of \citet{gloaguen2025dlmwm}. The exact
chainwash rates may differ for other DLMs, watermark designs, or decoding
settings. We evaluate four open-weight rewriters, but do not test closed-source
commercial humanizers or mixed-model chains where each hop uses a different
rewriter. Semantic preservation is measured using one sentence-encoder based
metric, so future work should compare multiple semantic metrics and human
judgments. We also do not study false attribution, where rewritten text is
tested against a different watermark key.



\end{document}